%% file: ijcai25.tex
\documentclass{article}
\usepackage{ijcai25}

\usepackage{times}
\usepackage{soul}
\usepackage{url}
\usepackage[hidelinks]{hyperref}
\usepackage[utf8]{inputenc}
\usepackage[small]{caption}
\usepackage{graphicx}
\usepackage{amsmath}
\usepackage{amsthm}
\usepackage{booktabs}
\usepackage{algorithm}
\usepackage{algorithmic}
\usepackage[switch]{lineno}

\usepackage{amsfonts}       
\usepackage{nicefrac}       
\usepackage{microtype}      
\usepackage{booktabs}       
\usepackage{xcolor}         
\usepackage{colortbl}
\usepackage{multirow}
\usepackage{makecell}
\usepackage{amsmath}
\usepackage{bm}

\title{Brain-Inspired Stepwise Patch Merging for Vision Transformers}

\author{
Yonghao Yu$^{1,2,3}$
\and
Dongcheng Zhao$^{2,3,5}$\and
Guobin Shen$^{2,3,4}$\and
Yiting Dong$^{2,3,4}$\And
Yi Zeng$^{1,2,3,4,5}$\footnote{Corresponding author.}\\
\affiliations
$^1$School of Artificial Intelligence, University of Chinese Academy of Sciences\\
$^2$Brain-inspired Cognitive AI Lab, Institute of Automation, Chinese Academy of Sciences\\
$^3$State Key Laboratory of Brain Cognition and Brain-inspired Intelligence Technology\\
$^4$School of Future Technology, University of Chinese Academy of Sciences\\
$^5$Center for Long-term AI\\
\emails
\{yuyonghao2023, zhaodongcheng2016, shenguobin2021, dongyiting2020, yi.zeng\}@ia.ac.cn
}

\begin{document}

\maketitle

\input{src/0_Abstract}

\input{src/1_Introduction}

\input{src/2_Related_Work}

\input{src/3_Methodology}

\input{src/4_Experiments}

\input{src/5_Ablation_Study}

\input{src/6_Future_Work}

\input{src/7_Conclusion}

\section*{Acknowledgments}
This work is supported by the National Natural Science Foundation of China (Grant No. 62406325).

\bibliographystyle{named}
\bibliography{ijcai25}

\end{document}

%% file: src/0_Abstract.tex
\begin{abstract}

The hierarchical architecture has become a mainstream design paradigm for Vision Transformers (ViTs), with Patch Merging serving as the pivotal component that transforms a columnar architecture into a hierarchical one.
Drawing inspiration from the brain's ability to integrate global and local information for comprehensive visual understanding, we propose Stepwise Patch Merging (SPM), which enhances the subsequent attention mechanism's ability to 'see' better.
SPM consists of Multi-Scale Aggregation (MSA) and Guided Local Enhancement (GLE) striking a proper balance between long-range dependency modeling and local feature enhancement.
Extensive experiments conducted on benchmark datasets, including ImageNet-1K, COCO, and ADE20K, demonstrate that SPM significantly improves the performance of various models, particularly in dense prediction tasks such as object detection and semantic segmentation.
Meanwhile, experiments show that combining SPM with different backbones can further improve performance.
The code has been released at \url{https://github.com/Yonghao-Yu/StepwisePatchMerging}.

\end{abstract}

%% file: src/1_Introduction.tex
\section{Introduction}

Transformers have demonstrated remarkable advancements in natural language processing (NLP)~\cite{vaswani2017attention,devlin2018bert}, and their application has recently extended significantly into the computer vision (CV) domain~\cite{dosovitskiy2020image}.
To enhance their adaptability to downstream tasks, hierarchical vision transformers (HVTs)~\cite{wang2021pyramid,liu2021swin} have been developed.
These architectures draw inspiration from the pyramid structure utilized in convolutional neural networks (CNNs)~\cite{krizhevsky2012imagenet,he2016deep}.
In HVTs, transformer blocks are segmented into multiple stages, resulting in a progressive reduction of feature map sizes and an increase in the number of channels as the network depth increases.

\begin{figure}[t]
    \centering
    \includegraphics[width=\columnwidth]{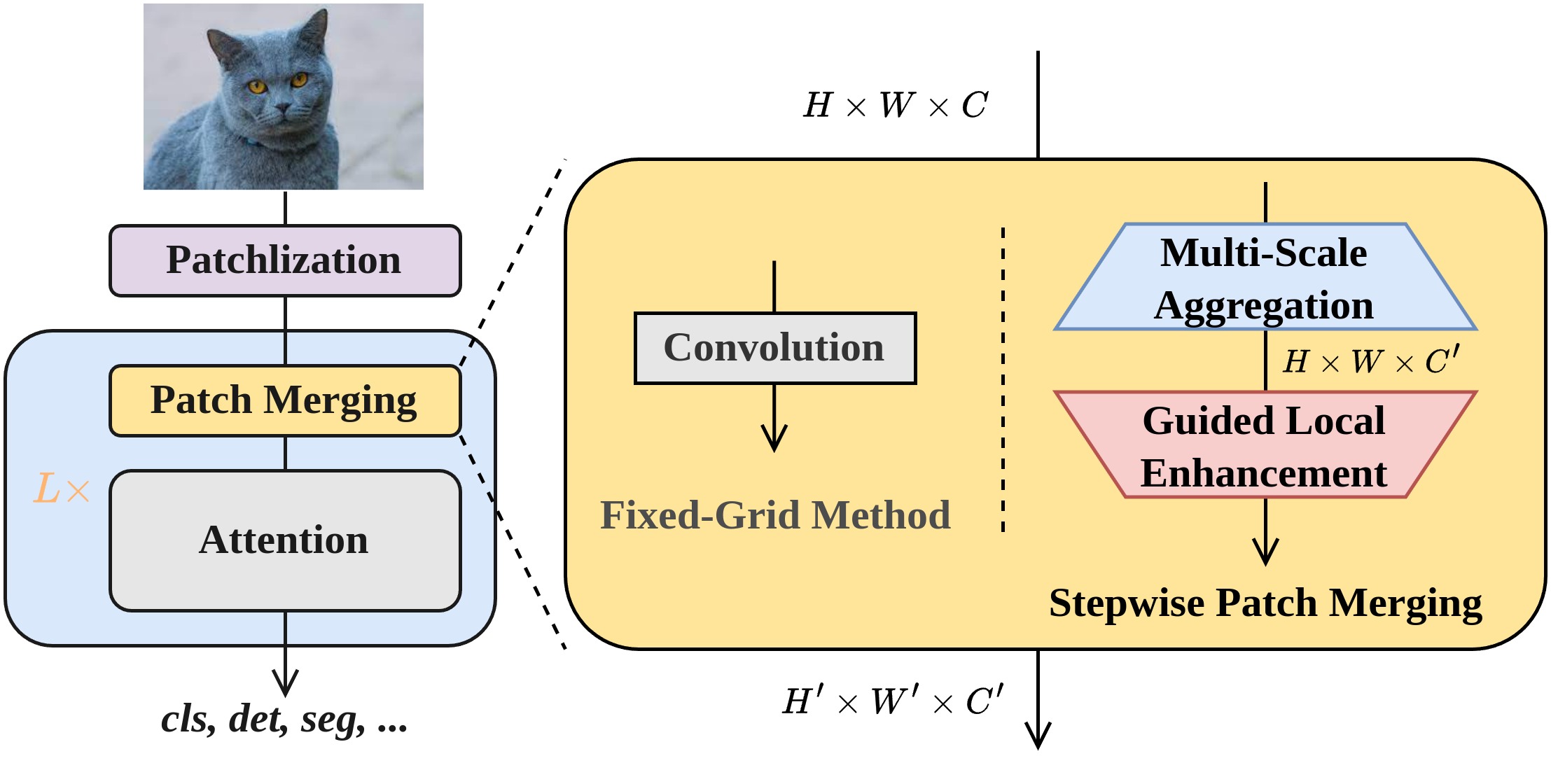}
    \caption{
   Overview of the proposed Stepwise Patch Merging (SPM) framework built upon the HVTs.
   The SPM framework comprises two sequential modules: Multi-Scale Aggregation (MSA) and Guided Local Enhancement (GLE).
      }
    \label{fig:spm}
\end{figure}

HVTs commonly utilize either standard convolutional layers or linear projection layers to amalgamate adjacent tokens~\cite{wang2021pyramid,liu2021swin,ren2022shunted}, with the objective of generating hierarchical feature maps.
Nevertheless, fixed-grid methods can limit the representational capacity of vision transformers in modeling geometric transformations, as not every pixel contributes equally to an output unit~\cite{luo2016understanding}.
To overcome this limitation, adaptive methods have been proposed to derive more informative downsampled tokens for subsequent processing.
For example, LIT~\cite{pan2022less}, drawing inspiration from deformable convolutions~\cite{dai2017deformable}, learns a grid of offsets to adaptively adjust spatial sampling locations for merging neighboring patches from a sub-window in a feature map.
Similarly, TCformer~\cite{zeng2022not} employs a variant of the k-nearest neighbor-based density peaks clustering algorithm (DPC-KNN)~\cite{du2016study} to aggregate redundant patches, generating more patches on the target object to capture additional information.
HAFA~\cite{chen2023building} integrates the methodologies of LIT and TCformer, predicting offsets to adjust the sampling center of patches in the shallow layers, while employing clustering in the deeper layers to group patches with similar semantics in the feature space.
However, these methods face several common challenges.
They often exhibit limited capacity for modeling long-distance relationships and suffer from a loss of spatial information due to the clustering process.
Additionally, the clustering algorithms used are typically not amenable to end-to-end training, leading to inefficiencies.
The integration of multiple modules, as seen in HAFA, further complicates their generalizability across different applications.

In the brain, the primary visual cortex (V1), integral to initial visual processing, houses neurons with relatively small receptive fields that are crucial for detecting fine, localized visual features such as edges and orientations.
As visual information propagates to higher cortical areas like V2, V3, and V4, neurons with increasingly larger receptive fields integrate these initial perceptions, facilitating the recognition of more complex patterns and broader contextual elements~\cite{hubel1962receptive}.
Additionally, the visual cortex benefits from a dynamic feedback system where higher-order areas like the inferotemporal cortex (IT) provide contextual modulation to lower areas.
This top-down modulation is essential for refining the perception of local features within their broader environmental matrix, enhancing both the accuracy and relevance of visual processing~\cite{gilbert2013top}.

Inspired by the orchestrated activities across various cortical areas, we introduce Stepwise Patch Merging (SPM), as shown in Fig.~\ref{fig:spm}, a novel approach designed to enhance the receptive field while preserving local details.
SPM framework consists of two sequential stages: Multi-Scale Aggregation (MSA) and Guided Local Enhancement (GLE).
In the MSA stage, spatial dimensions are preserved while channel dimensions are increased to a designated size.
This process aggregates multi-scale information, enriching the semantic content to accommodate the increased capacity of the feature map.
Subsequently, the GLE stage reduces the spatial dimensions of the feature map while maintaining the channel dimensions.
Given that the input to GLE already contains rich semantic information, this stage emphasizes local information, optimizing it for downstream dense prediction tasks such as object detection and semantic segmentation.
The distinct focus and reasonable division of labor between the MSA and GLE modules ensure that the SPM architecture serves as a flexible, drop-in replacement for existing HVTs.

In summary, our contributions are as follows:
\begin{itemize}
\item We propose an innovative technique termed Stepwise Patch Merging (SPM), which serves as a plug-in replacement within HVTs, leading to substantial performance enhancements.
\item The SPM framework comprises two distinct modules: Multi-Scale Aggregation (MSA) and Guided Local Enhancement (GLE). MSA enriches feature representation by integrating multi-scale information, while GLE enhances the extraction of local details, achieving an optimal balance between long-range dependency modeling and local feature refinement.
\item Extensive experiments conducted on benchmark datasets, including ImageNet-1K, COCO, and ADE20K, demonstrate that SPM significantly boosts the performance of various models, particularly in downstream dense prediction tasks such as object detection and semantic segmentation.
\end{itemize}

%% file: src/2_Related_Work.tex
\section{Related Work}

\subsection{Vision Transformer}

The Vision Transformer (ViT)~\cite{dosovitskiy2020image} revolutionized visual tasks by introducing the transformer architecture to computer vision.
ViT segments images into non-overlapping patches, projects these patches linearly into token sequences, and processes them using a transformer encoder.
ViT models have demonstrated superior performance in image classification and other downstream tasks, surpassing CNNs~\cite{he2016deep,krizhevsky2012imagenet} when trained with large-scale pretraining datasets and advanced training methodologies.
Motivated by the success of CNNs and the necessity to address dense prediction tasks, researchers have incorporated the feature pyramid structure within transformers.
This innovation has led to the development and widespread adoption of HVTs~\cite{dong2022cswin,ren2022shunted,guo2022cmt,hou2022conv2former,lin2023scale}.

\subsection{Hierarchical Feature Representation}

Hierarchical feature representation plays a pivotal role in dense prediction tasks, prompting extensive research in this domain.
Existing approaches can be broadly categorized into fixed-grid and dynamic feature-based methods.
Fixed-grid methods, exemplified by works such as PVT~\cite{wang2021pyramid} and Swin~\cite{liu2021swin}, merge patches within adjacent windows using 2D convolution.
In contrast, dynamic methods, such as DynamicViT~\cite{rao2021dynamicvit} adaptively extract features by eliminating redundant patches and retaining essential ones, thereby forming hierarchical feature maps.
EviT~\cite{liang2022not} enhances this approach by selecting the top K tokens with the highest average values across all heads for the next stage, merging the remaining tokens.
PS-ViT~\cite{yue2021vision} further refines the process by iteratively adjusting patch centers towards the object to enrich object information within the hierarchical feature maps.
Token Merging~\cite{bolya2022token} employs cosine similarity to progressively merge similar tokens, thereby increasing model throughput.

Fixed-grid methods are constrained by their singular and relatively small receptive fields, and excessively enlarging the grid size leads to increased computational overhead.
Dynamic feature-based methods, while adaptive, may discard low-scoring tokens that contain valuable information and often lack end-to-end training capabilities.
Our proposed SPM distinguishes itself from both fixed-grid and dynamic feature-based methods.

\subsection{Global/Coarse and Local/Fine Feature Fusion}

Fusing coarse- and fine-grained features is not novel; however, most prior work treats both feature types equally and performs simple fusion.
For example, PLG-ViT~\cite{ebert2023plg} fuses global and local features using the Hadamard product.
SPM uses a lightweight module (GTG) to generate global information that guides the generation of detailed features, an approach demonstrated by experiments to be highly effective.

%% file: src/3_Methodology.tex
\section{Methodology}

Inspired by the brain's ability to integrate global and local information when processing visual scenes, we propose the Stepwise Patch Merging framework, as illustrated in Fig.~\ref{fig:spm}.
The framework comprises two primary components: Multi-Scale Aggregation (MSA) and Guided Local Enhancement (GLE), designed to address variations in feature map dimensions.
The MSA module enhances feature diversity by increasing the number of channels and capturing long-range dependencies, akin to how the brain processes information at multiple scales to form a coherent perception.
In contrast, the GLE module optimizes local feature extraction by introducing context-aware guide tokens within local windows, thereby refining and enhancing feature details.
This synergistic design effectively combines the strengths of both global structure processing and local detail enhancement, making it particularly beneficial for downstream dense prediction tasks.

\begin{figure}[t]
    \centering
    \includegraphics[width=\columnwidth]{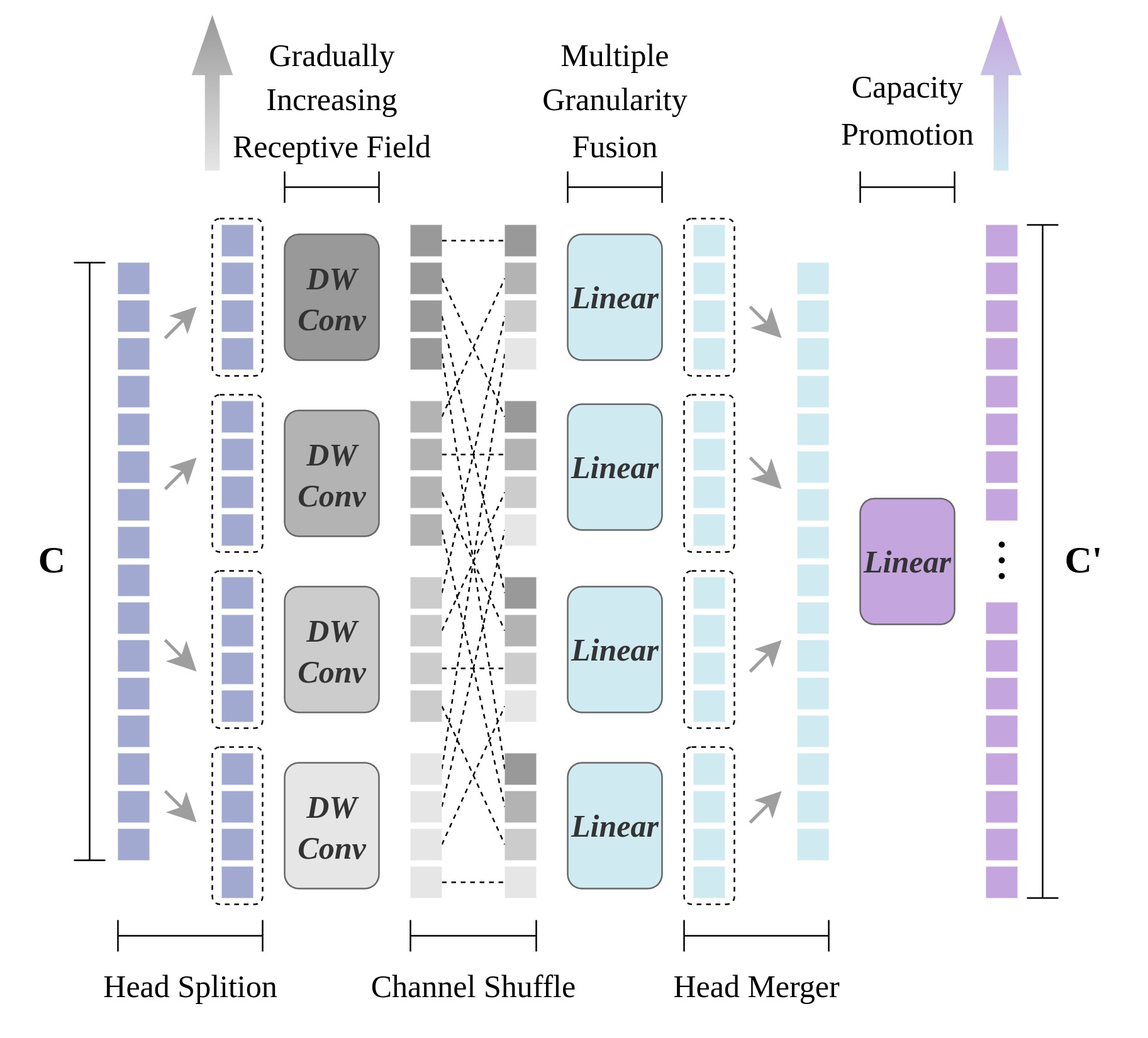}
    \caption{An illustration of Multi-Scale Aggregation (MSA).}
    \label{fig:step1}
\end{figure}

\subsection{Multi-Scale Aggregation}

Our proposed Multi-Scale Aggregation (MSA) module draws inspiration from the brain's remarkable ability to effectively model long-range dependencies when processing visual information.
In the brain, the visual system achieves precise modeling of long-range dependencies through multi-level and multi-scale information processing.
Neurons with small receptive fields process local features, and this information is progressively integrated over larger areas by neurons with larger receptive fields, capturing complex patterns and objects.
Additionally, the brain's extensive network of long-range neural connections allows for the exchange and integration of data from various parts of the visual field, facilitating a comprehensive understanding of the scene.
Furthermore, neurons within the same level possess receptive fields of varying sizes, enabling the brain to simultaneously process local details and global features.
This sophisticated mechanism of combining local and global information processing in the brain inspired the design of our MSA module, which aims to enhance feature diversity and capture long-range dependencies effectively.

Inspired by these mechanisms, the MSA module first divides the input channels $C$ into $N$ distinct heads, each undergoing depth-wise convolutions with varying receptive fields.
This method not only reduces the parameter count and computational cost but also facilitates the extraction of multi-granularity information, akin to how different neurons in the brain handle information processing.
Subsequently, the MSA module employs larger convolutional kernels to further expand the receptive field, thereby enhancing its capability to model long-range dependencies.
Following this, Channel Shuffle \cite{zhang2018shufflenet} is used to interleave channels containing features of different scales, followed by a series of linear projections to fuse these multi-scale features.
The number of linear projections is $\frac{C}{N}$, with each projection having unique parameters.
Finally, the $N$ heads are concatenated, and a final linear projection adjusts the number of channels to the specified $C^{\prime}$.

By leveraging the brain's mechanism for effective long-range dependency modeling, the MSA module better captures and integrates key features, significantly enhancing the model's performance in complex visual recognition tasks.

Our proposed MSA can be formulated as follows:
\begin{equation}
    \begin{aligned}
        &H_n = DWConv_{k_n\times k_n}(x_n) \\
        &G^{c} = \mathbf{W}^{c}([H_1^{c}; H_2^{c}; ...; H_N^{c}]) \\
        &\bm{MSA}(X) = \mathbf{W}([G^1; G^2; ...; G^{\frac{C}{N}}])
    \end{aligned}
\end{equation}
where $X=[x_1, x_2, \ldots , x_N]$ represents the input $X$ split into multiple heads along the channel dimension, and $x_n \in \mathbb{R}^{B \times H \times W \times \frac{C}{N}}$ denotes the $n$-th head. The kernel size of the depth-wise convolution for the $n$-th head is denoted by $k_n \in {k_1, k_2, \ldots, k_N}$. Here, $H_n \in \mathbb{R}^{B \times H \times W \times \frac{C}{N}}$ represents the $n$-th head after being processed by the depth-wise convolution with $out_channels = in_channels$, and $H_n^{c}$ represents the $c$-th channel in the $n$-th head. $\mathbf{W}^{c} \in \mathbb{R}^{N \times N}$ is the weight matrix of the linear projection. $\mathbf{G}^{c} \in \mathbb{R}^{B \times H \times W \times N}$. Finally, $\mathbf{W} \in \mathbb{R}^{C \times C^{\prime}}$ is the weight matrix of the linear projection that adjusts the number of channels to the specified $C^{\prime}$.

\subsection{Guided Local Enhancement}

Inspired by the brain's ability to enhance local features through context-aware processing, we developed the Guided Local Enhancement (GLE) module.
In the brain, local feature enhancement is achieved by integrating information from both local and global contexts.
Higher-level cortical areas provide contextual feedback that refines the processing of local features, ensuring that details are interpreted within the broader visual context.
This hierarchical processing involves neurons that respond specifically to local stimuli but are influenced by surrounding contextual information, allowing for more nuanced and precise feature extraction.

\begin{figure}[t]
    \centering
    \includegraphics[width=\columnwidth]{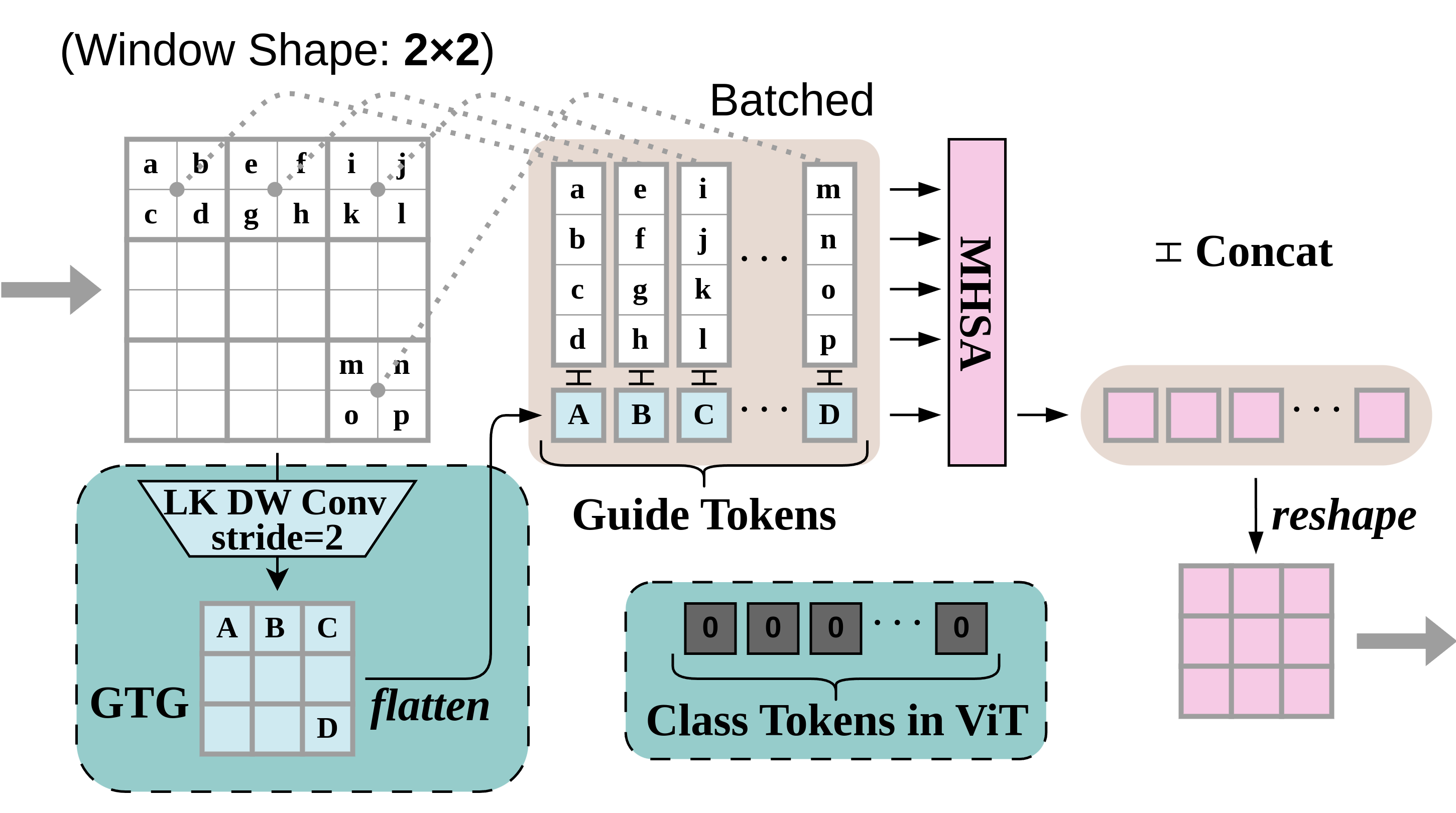}
    \caption{An illustration of Guided Local Enhancement (GLE).}
    \label{fig:step2}
\end{figure}

Following this principle, the GLE module acts as a local feature enhancer utilizing context-aware guide tokens, as illustrated in Fig.~\ref{fig:step2}.
Specifically, we implement self-attention within a local window and introduce a $[Guide]$ token into the input space.
This $[Guide]$ token undergoes the same self-attention operations as the patch tokens, thereby providing high-level semantic features to the local window during pooling.
By mimicking the brain's method of using contextual information to refine local feature extraction, the GLE module ensures that the extracted local features are both precise and contextually relevant.

Formally, given an input $X \in \mathbb{R}^{C \times H \times W}$, the $[Guide]$ tokens are generated by a large-kernel depth-wise convolution, referred to as the Guide Token Generator (GTG), which can be described as follows:

\begin{equation}
    \begin{aligned}
        \bm{GTG}(X) = DWConv(GELU(BatchNorm(X))).
    \end{aligned}
\end{equation}

We focus on the operations performed on a single pixel within the input feature map.
We define a set of pixels within a local window centered at pixel $(i, j)$ as $\rho(i, j)$.
For a fixed window size of $k \times k$, $\Vert\rho(i,j)\Vert = k^2$.
In our setup, $k$ is equal to the stride of the GTG, both being 2, meaning $\#Windows = \Vert GTG(X)\Vert = \frac{H\times W}{4}$.
Tokens within a window containing a $[Guide]$ token can be represented by the sequence $\mathbf{z}$:
\begin{equation}
    \begin{aligned}
        \mathbf{z} = \left[S_{(i,j)\sim GTG(X)}; S^1_{(i,j)\sim \rho(i,j)}; \cdots; S^{k^2}_{(i,j)\sim \rho(i,j)}\right],
    \end{aligned}
\end{equation}
then perform the standard self-attention operation on $\mathbf{z}$:
\begin{equation}
    \begin{aligned}
        \left[\mathbf{q}, \mathbf{k}, \mathbf{v}\right] &= \mathbf{z}\mathbf{U}_{qkv}, \\
        SA(\mathbf{z}) &= softmax(\frac{\mathbf{q}\mathbf{k}^\top}{\sqrt{C}})\mathbf{v}, \\
    \end{aligned}
\end{equation}
where $\mathbf{U}_{qkv}\in \mathbb{R} ^ {C\times 3C}$, and we ignore the relative positional relationships between tokens within a window, so positional encoding is not used.
Finally, we select the $[Guide]$ token as the output of the GLE:
\begin{equation}
    \begin{aligned}
        &SA(\mathbf{z}) = \left[A_{(i,j)\sim GTG(X)}; A_{(i,j)\sim \rho(i,j)}\right],\\
        &\bm{GLE}(X_{(i,j)}) = A_{(i,j)\sim GTG(X)}.
    \end{aligned}
\end{equation}

It is worth noting that when the $[Guide]$ token is stripped of its semantic information, it degrades into the $[CLS]$ token in the vanilla vision transformer \cite{dosovitskiy2020image}. Experiments show that using our $[Guide]$ tokens results in higher performance (see Tab.~\ref{table:clstoken}).

%% file: src/4_Experiments.tex
\section{Experiments}

\subsection{Image Classification on ImageNet-1K}

\paragraph{Setting.}
We first evaluate the proposed SPM framework on the ImageNet-1K dataset~\cite{deng2009imagenet}, which comprises 1.28 million training images and 50,000 validation images spanning 1,000 categories.
To ensure a fair comparison, all models are trained on the training set and report the top-1 error rate on the validation set.
For data augmentation, we apply a suite of techniques including random cropping, random horizontal flipping~\cite{szegedy2015going}, label smoothing regularization~\cite{szegedy2016rethinking}, mixup~\cite{zhang2017mixup}, CutMix~\cite{yun2019cutmix}, and random erasing~\cite{zhong2020random}.
These augmentations are employed to enhance the robustness and generalization ability of the models.
During training, we use the AdamW optimizer~\cite{loshchilov2017decoupled} with a momentum parameter of 0.9, a mini-batch size of 128, and a weight decay of $5\times 10^{-2}$.
The initial learning rate is set to $1\times 10^{-3}$ and follows a cosine annealing schedule~\cite{loshchilov2016sgdr} to gradually reduce the learning rate.
All models are trained from scratch for 300 epochs on eight NVIDIA A100 GPUs.
For evaluation, we adopt the standard center crop strategy on the validation set, where a $224\times 224$ patch is extracted from each image to assess the classification accuracy.

\begin{table}[H]
    \centering
    \resizebox{\columnwidth}{!}{
    \begin{tabular}{c|c|c}
        \toprule
        Backbone                                         & \makecell[c]{\#Params \\ (M)} & \makecell[c]{Top-1 Acc. \\ (\%)}      \\
        \midrule
        ResNet18~\cite{he2016deep}                       & 11.7                          & 68.5                                  \\
        DeiT-Tiny/16~\cite{touvron2021training}          & 5.7                           & 72.2                                  \\
        PVT-Tiny~\cite{wang2021pyramid}                  & 13.2                          & 75.1                                  \\
        PVT-Tiny~(HAFA)~\cite{chen2023building}          & 14.6                          & 77.5                                  \\
        \rowcolor{gray!20}\textbf{PVT-Tiny~(SPM)}        & 14.0                          & \textbf{79.5 \textcolor{red}{(+4.4)}} \\
        \midrule
        ResNet50~\cite{he2016deep}                       & 25.6                          & 78.5                                  \\
        ResNeXt50-32$\times$4d~\cite{xie2017aggregated}  & 25.0                          & 79.5                                  \\
        DeiT-Small/16~\cite{touvron2021training}         & 22.1                          & 79.9                                  \\
        HRNet-W32~\cite{wang2020deep}                    & 41.2                          & 78.5                                  \\
        PVT-Small~\cite{wang2021pyramid}                 & 24.5                          & 79.8                                  \\
        PVT-Small~(HAFA)~\cite{chen2023building}         & 25.8                          & 80.1                                  \\
        \rowcolor{gray!20}\textbf{PVT-Small~(SPM)}       & 25.3                          & \textbf{81.7 \textcolor{red}{(+1.9)}} \\
        \midrule
        ResNeXt101-64$\times$4d~\cite{xie2017aggregated} & 83.5                          & 81.5                                  \\
        ViT-Base/16~\cite{dosovitskiy2020image}          & 86.6                          & 81.8                                  \\
        DeiT-Base/16~\cite{touvron2021training}          & 86.6                          & 81.8                                  \\
        PVT-Medium~\cite{wang2021pyramid}                & 44.2                          & 81.2                                  \\
        \rowcolor{gray!20}\textbf{PVT-Medium~(SPM)}      & 45.0                          & \textbf{81.9 \textcolor{red}{(+0.7)}} \\
        \bottomrule
    \end{tabular}}
    \caption{Image classification performance on the ImageNet validation set. “\#Params” refers to the number of parameters.}
    \label{table:ImageNet}
\end{table}

\begin{table*}[htbp]
    \centering
    \resizebox{\textwidth}{!}{
    \begin{tabular}{c|c|cccccc|ccc}
        \toprule
        \multirow{2}*{Backbone}                          & \multirow{2}*{\makecell[c]{\#Params \\ (M)}} & \multicolumn{9}{c}{Mask R-CNN} \\
        \cline{3-11}
        ~                                                & ~ & $AP^{b}$ & $AP_{50}^{b}$ & $AP_{75}^{b}$ & $AP_{s}^{b}$ & $AP_{m}^{b}$ & $AP_{l}^{b}$ & $AP^{m}$ & $AP_{50}^{m}$ & $AP_{75}^{m}$ \\
        \midrule
        ResNet18~\cite{he2016deep}                       & 31.2 & 34.0 & 54.0 & 36.7 & - & - & - & 31.2 & 51.0 & 32.7\\
        PVT-Tiny~\cite{wang2021pyramid}                  & 32.9 & 36.7 & 59.2 & 39.3 & 21.6 & 39.2 & 49.0 & 35.1 & 56.7 & 37.3 \\
        PVT-Tiny~(HAFA)~\cite{chen2023building}          & 34.5 & 39.8 & 62.6 & 43.3 & 23.3 & 42.7 & 53.3 & 37.1 & 59.4 & 39.3 \\
        \rowcolor{gray!20}\textbf{PVT-Tiny~(SPM)}        & 33.7 & \textbf{40.8 \textcolor{red}{(+4.1)}} & \textbf{63.4} & \textbf{44.3} & \textbf{24.9} & \textbf{44.0} & \textbf{54.0} & \textbf{38.0 \textcolor{red}{(+2.9)}} & \textbf{60.4} & \textbf{40.6} \\
        \midrule
        ResNet50~\cite{he2016deep}                       & 44.2 & 38.0 & 58.6 & 41.4 & - & - & - & 34.4 & 55.1 & 36.7\\
        PVT-Small~\cite{wang2021pyramid}                 & 44.1 & 40.4 & 62.9 & 43.8 & 22.9 & 43.0 & 55.4 & 37.8 & 60.1 & 40.3 \\
        PVT-Tiny~(HAFA)~\cite{chen2023building}          & 45.8 & 41.8 & 64.4 & 45.7 & 26.0 & 44.6 & 56.1 & 38.9 & 61.5 & 41.9 \\
        \rowcolor{gray!20}\textbf{PVT-Small~(SPM)}       & 44.9 & \textbf{43.0 \textcolor{red}{(+2.6)}} & \textbf{65.4} & \textbf{46.7} & \textbf{25.5} & \textbf{46.2} & \textbf{57.6} & \textbf{39.6 \textcolor{red}{(+1.8)}} & \textbf{62.3} & \textbf{42.4} \\
        \midrule
        ResNet101~\cite{he2016deep}                      & 63.2 & 40.4 & 61.1 & 44.2 & - & - & - & 36.4 & 57.7 & 38.8 \\
        ResNeXt101-32$\times$4d~\cite{xie2017aggregated} & 62.8 & 41.9 & 62.5 & 45.9 & - & - & - & 37.5 & 59.4 & 40.2 \\
        PVT-Medium~\cite{wang2021pyramid}                & 63.9 & 42.0 & 64.4 & 45.6 & - & - & - & 39.0 & 61.6 & 42.1 \\
        \rowcolor{gray!20}\textbf{PVT-Medium~(SPM)}      & 64.7 & \textbf{43.3 \textcolor{red}{(+1.3)}} & \textbf{64.9} & \textbf{47.6} & \textbf{25.8} & \textbf{46.4} & \textbf{58.3} & \textbf{39.4 \textcolor{red}{(+0.4)}} & \textbf{61.7} & \textbf{42.2} \\
        \bottomrule
    \end{tabular}}
    \caption{Object detection and instance segmentation performance on COCO val2017. $AP^{b}$ and $AP^{m}$ denote bounding box AP and mask AP, respectively.}
    \label{table:COCO}
\end{table*}

\paragraph{Result.}
In Tab.~\ref{table:ImageNet}, we observe that incorporating the SPM framework into the PVT results in significant improvements in classification accuracy, specifically by 4.4\%, 1.9\%, and 0.7\% in the Tiny, Small, and Medium models, respectively.
The final experimental results indicate that the accuracy of models of various sizes has been enhanced, with the most notable improvement observed in the Tiny model.
Furthermore, with the integration of SPM, PVT-Small surpassed PVT-Medium by 0.5\%.

\subsection{Object Detection on COCO}

\paragraph{Setting.}
Object detection and instance segmentation experiments were conducted on the challenging COCO benchmark~\cite{lin2014microsoft}.
All models were trained on the training set comprising 118k images and evaluated on the validation set with 5k images.
We validated the effectiveness of different backbones using Mask R-CNN~\cite{he2017mask}.
Before training, the weights pre-trained on ImageNet-1K were used to initialize the backbone, and the newly added layers were initialized using the Xavier initialization method~\cite{glorot2010understanding}.
Our models were trained with a batch size of 16 on 8 NVIDIA A100 GPUs and optimized using the AdamW optimizer~\cite{loshchilov2017decoupled} with an initial learning rate of $1\times 10^{-4}$.

\paragraph{Result.}
As shown in Tab.~\ref{table:COCO}, incorporating the SPM framework into the PVT resulted in significant improvements of 4.1\%, 2.6\%, and 1.3\% in the Tiny, Small, and Medium models, respectively, for the object detection task.
Models integrated with SPM show marked improvements in detecting medium-sized and large objects.
This enhancement is attributed to the original patch merging's relatively singular and small receptive fields, whereas the MSA module integrates features with diverse receptive fields, enabling the model to more accurately capture long-range dependencies.
Moreover, there is a significant improvement in detecting small objects.
Although larger models are typically better at modeling global relationships, the disruption of local information may hinder small objects from establishing complete semantic information, leading to missed detections.
The GLE module addresses this by enhancing the perception of local discriminative information, resulting in consistent improvements in the detection performance of small objects across models of different sizes.

\subsection{Semantic Segmentation on ADE20K}

\begin{table}[htbp]
    \centering
    \resizebox{\columnwidth}{!}{
    \begin{tabular}{c|c|c}
        \toprule
        \multirow{2}*{Backbone}                          & \multicolumn{2}{c}{Semantic FPN}                     \\
        \cline{2-3}
        ~                                                & \#Params (M) & mIoU (\%)                             \\
        \midrule
        ResNet18~\cite{he2016deep}                       & 15.5         & 32.9                                  \\
        PVT-Tiny~\cite{wang2021pyramid}                  & 17.0         & 35.7                                  \\
        PVT-Tiny~(HAFA)~\cite{chen2023building}          & 18.7         & 40.1                                  \\
        \rowcolor{gray!20}\textbf{PVT-Tiny~(SPM)}        & 17.8         & \textbf{41.5 \textcolor{red}{(+5.8)}} \\
        \midrule
        ResNet50~\cite{he2016deep}                       & 28.5         & 36.7                                  \\
        PVT-Small~\cite{wang2021pyramid}                 & 28.2         & 39.8                                  \\
        PVT-Small~(HAFA)~\cite{chen2023building}         & 29.9         & 43.8                                  \\
        \rowcolor{gray!20}\textbf{PVT-Small~(SPM)}       & 29.0         & \textbf{45.9 \textcolor{red}{(+6.1)}} \\
        \midrule
        ResNet101~\cite{he2016deep}                      & 47.5         & 38.8                                  \\
        ResNeXt101-32$\times$4d~\cite{xie2017aggregated} & 47.1         & 39.7                                  \\
        PVT-Medium~\cite{wang2021pyramid}                & 48.0         & 41.6                                  \\
        \rowcolor{gray!20}\textbf{PVT-Medium~(SPM)}      & 48.8         & \textbf{45.3 \textcolor{red}{(+3.7)}} \\
        \bottomrule
    \end{tabular}}
    \caption{Semantic segmentation performance of different backbones on the ADE20K validation set.}
    \label{table:ADE20K}
\end{table}

\paragraph{Setting.}
The ADE20K dataset~\cite{zhou2017scene} is a widely utilized benchmark for semantic segmentation, comprising 150 categories with 20,210 images for training, 2,000 images for validation, and 3,352 images for testing.
All the methods compared were evaluated using the Semantic FPN framework~\cite{kirillov2019panoptic}.
The backbone network of our method was initialized with the pre-trained ImageNet-1k model, and the newly added layers were initialized using the Xavier initialization method.
The initial learning rate was set to 0.0001, and the model was optimized using the AdamW optimizer.
We trained our models for 40,000 iterations with a batch size of 16 on eight NVIDIA A100 GPUs.
The learning rate followed a polynomial decay schedule with a power of 0.9.
During training, images were randomly resized and cropped to $512\times 512$ pixels.
For testing, images were rescaled to have a shorter side of 512 pixels.

\paragraph{Result.}
As shown in Tab.~\ref{table:ADE20K}, the integration of the SPM framework led to a significant enhancement in the semantic segmentation task.
Specifically, the performance of the Tiny, Small, and Large models improved by 5.8\%, 6.1\%, and 3.7\%, respectively.
Interestingly, in both classification and detection tasks, the relative improvement brought by SPM compared to the base model gradually decreases as the model size increases.
However, on the ADE20K dataset, the performance gain of PVT-Small exceeds that of PVT-Tiny, with improvements of 6.1\% and 5.8\%, respectively.

\begin{figure}[t]
    \centering
    \includegraphics[width=\columnwidth]{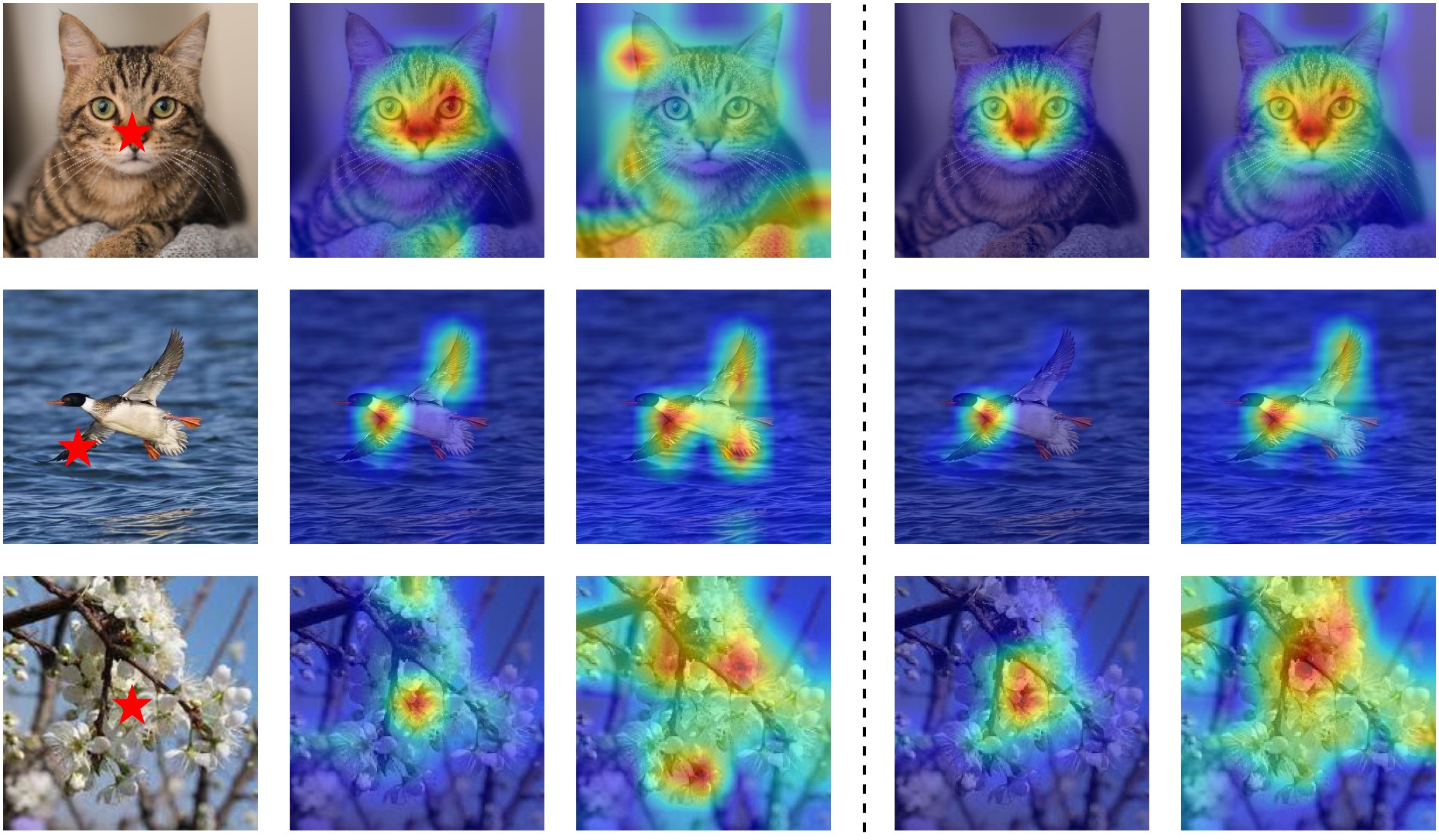}
    \caption{
    Visualization of the attention map includes the original images (the 1st column), the visualization of the attention map for each head after using SPM (the 2nd and 3rd columns), and without using SPM (the 4th and 5th columns).
    The red five-pointed star in the original images represents the position of the query.
    }
    \label{fig:attn_vis}
\end{figure}

\subsection{Effectiveness on Other Backbones}

To further validate the generalizability of the proposed SPM framework, we integrated SPM into various mainstream Transformer backbones and trained them on the ImageNet-1K dataset.
We employed consistent training settings to ensure a fair comparison, and the top-1 accuracies are presented in Tab.~\ref{table:generalization}.
The results demonstrate that the performance improvements conferred by SPM are universal across different backbones, indicating its robust generalization capability.
Specifically, our method significantly enhanced the performance of Swin-T by 1.1\%, Shunted-T by 0.8\%, and NAT-Mini by 0.4\% on ImageNet-1K.
These results underscore the effectiveness of SPM in boosting the performance of various Transformer architectures, highlighting its potential as a versatile enhancement technique in the field of computer vision.

\subsection{Visualization}

We compared the visualization results of the attention maps with and without using the SPM framework, as shown in Fig.~\ref{fig:attn_vis}.
Specifically, we replaced the original patch merging block of PVT-Tiny with our SPM and visualized the first block of the second stage for two separate heads.
For example, after employing SPM, the bird's two wings and tail were successfully linked, whereas the vanilla PVT-Tiny failed to capture the distant tail.
This demonstrates that SPM facilitates the network's ability to establish long-range relationships at shallower layers, leading to significant improvements in classification performance.

\begin{figure}[t]
    \centering
    \includegraphics[width=\columnwidth]{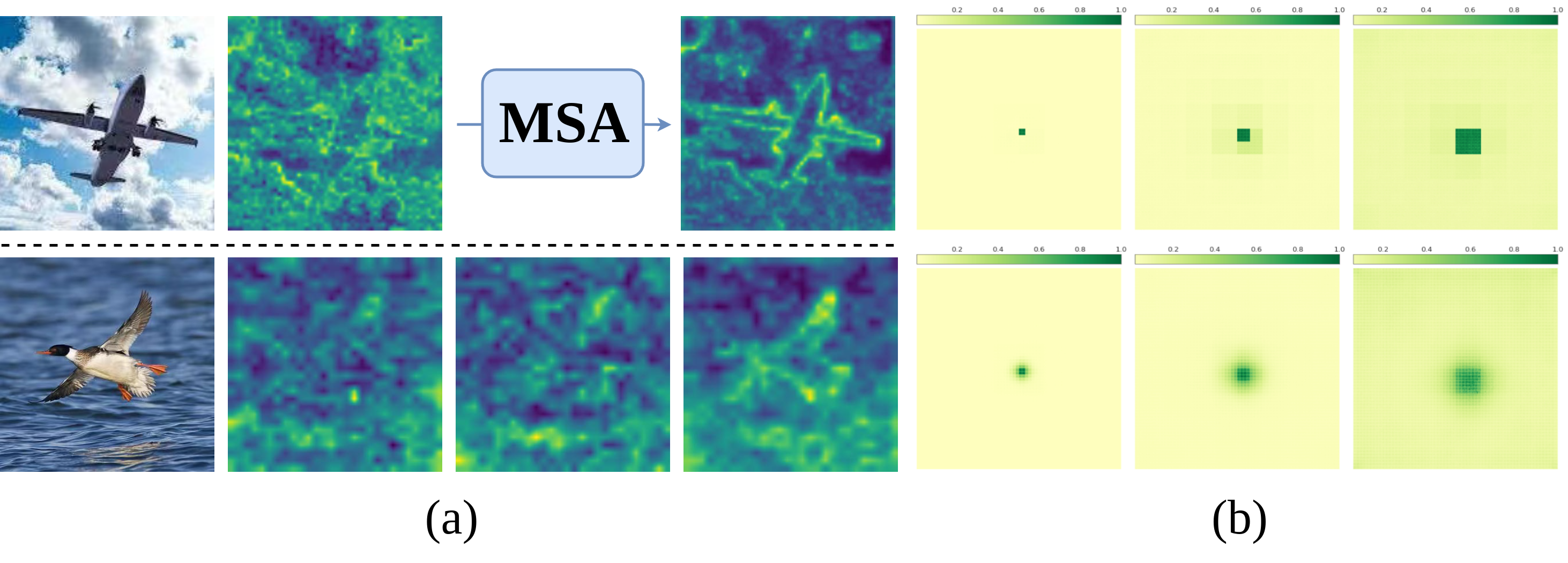}
    \caption{
    (a) Visualizations before/after MSA and GLE with increasing GTG kernel size (3, 5 and 7); 
    (b) Visualization comparison of the Effective Receptive Field of PVT-Tiny before (the 1st row) and after applying our SPM (the 2nd row), using the outputs from the three stages separately.
    Each ERF image is generated by averaging over 5000 $256\times 256$ images from ImageNet-1K validation set.
    }
    \label{visual_erf}
\end{figure}

As shown in Fig.~\ref{visual_erf}~(a), MSA enhances fine-grained features and moderate global context benefits detail generation, while excessive global aggregation leads to performance degradation (see Tab.~\ref{table:stgks}).

Additionally, we employed the Effective Receptive Field (ERF) method~\cite{luo2016understanding} as a visualization tool to compare the changes in ERF before and after using SPM, as depicted in Fig.~\ref{visual_erf}.
It is readily observed that after integrating SPM, the size of the ERF not only increases significantly but also changes shape from a regular square to a radial decay pattern, which aligns more closely with biological vision.

\begin{table}[htbp]
    \centering
    \resizebox{\columnwidth}{!}{
    \begin{tabular}{c|c|c|c}
        \toprule
        Reference                & Backbone                                    & \makecell[c]{\#Params \\ (M)} & \makecell[c]{Top-1 Acc. \\ (\%)}                         \\
        \midrule
        \multirow{3}*{ICCV 2021} & Swin-T~\cite{liu2021swin}                   & 29.0                          & 81.3                                                     \\
        ~                        & Swin-T~(HAFA)~\cite{liu2021swin}            & 29.1                          & 81.7                                                     \\
        \multicolumn{1}{c|}{~}   & \cellcolor{gray!20}\textbf{Swin-T~(SPM)}    & \cellcolor{gray!20}29.9       & \cellcolor{gray!20}\textbf{82.4 \textcolor{red}{(+1.1)}} \\
        \midrule
        \multirow{2}*{CVPR 2022} & Shunted-T~\cite{ren2022shunted}             & 11.5                          & 79.8                                                     \\
        \multicolumn{1}{c|}{~}   & \cellcolor{gray!20}\textbf{Shunted-T~(SPM)} & \cellcolor{gray!20}11.6       & \cellcolor{gray!20}\textbf{80.6 \textcolor{red}{(+0.8)}} \\
        \midrule
        \multirow{2}*{CVPR 2023} & NAT-Mini~\cite{hassani2023neighborhood}     & 20.0                          & 81.8                                                     \\
        \multicolumn{1}{c|}{~}   & \cellcolor{gray!20}\textbf{NAT-Mini~(SPM)}  & \cellcolor{gray!20}19.9       & \cellcolor{gray!20}\textbf{82.2 \textcolor{red}{(+0.4)}} \\
        \bottomrule
    \end{tabular}}
    \caption{SPM can boost backbones with different attention mechanisms via replacing their original Patch Merging blocks. “\#Params” refers to the number of parameters.}
    \label{table:generalization}
\end{table}

%% file: src/5_Ablation_Study.tex
\section{Ablation Study}

All experiments were conducted with the random seed fixed at 0, without tuning for desired outcomes.

\subsection{The Effectiveness of MSA, GLE and GTG}

As presented in Tab.~\ref{abla}, replacing GLE with a $3\times 3$ AvgPool results in a 2.4\% performance decrease, but still outperforms the baseline by 2.0\%, with nearly identical parameter count and computational complexity.

Although the DWConv in the GTG module introduces a small number of learnable parameters, it has been experimentally proven to be highly effective.
For instance, replacing GTG with an AvgPool of the same kernel size results in a 1.5\% performance drop.

Moreover, PVT-Tiny~(SPM) achieves accuracy similar to PVT-Small while using fewer FLOPs and parameters, suggesting that SPM's advantage comes from structural design rather than resource scaling.

\begin{table}[t]
    \centering
    \resizebox{\columnwidth}{!}{
    \begin{tabular}{c|c|c|c|c}
        \toprule
        Backbone                & Impl. of Patch Merging                      & \makecell[c]{\#P \\ (M)}         & \makecell[c]{\#F \\ (G)}        & \makecell[c]{Top-1 Acc. \\ (\%)} \\
        \midrule
        \multirow{6}*{PVT-Tiny} & \cellcolor{gray!20}\textbf{SPM (MSA + GLE)} & \cellcolor{gray!20}\textbf{13.6} & \cellcolor{gray!20}\textbf{2.7} & \cellcolor{gray!20}\textbf{79.5} \\
        ~                       & MSA $\rightarrow$ $3\times 3$ Conv.         & 15.1                             & 3.3                             & 78.8 (-0.7)                      \\
        ~                       & GLE $\rightarrow$ $2\times 2$ Conv.         & 14.0                             & 2.2                             & 77.2 (-2.3)                      \\
        ~                       & GLE $\rightarrow$ $3\times 3$ AvgPool       & 12.4                             & 2.0                             & 77.1 (-2.4)                      \\
        ~                       & GTG $\rightarrow$ $7\times 7$ AvgPool       & 13.6                             & 2.7                             & 78.0 (-1.5)                      \\
        ~                       & $2\times 2$ Conv.                           & 12.8                             & 1.9                             & 75.1                             \\
        \midrule
        PVT-Small               & $2\times 2$ Conv.                           & 24.1                             & 3.7                             & 79.8                             \\
        \bottomrule
    \end{tabular}}
    \caption{Classification results on ImageNet-1K with different patch merging strategies. “\#P” and “\#F” are calculated by the \textit{thop} package.}
    \label{abla}
\end{table}

As shown in Fig.~\ref{flops_loss}~(a), the additional 0.8 GFLOPs overhead mainly results from the linear projection in GLE, which fuses $[Guide]$ and regular tokens via self-attention.
Initially, we experimented with replacing self-attention with dynamic weighting (SKNet~\cite{li2019selective}), which reduced FLOPs but caused a slight drop in performance.
This highlights the trade-off between efficiency and accuracy, and supports our choice of maintaining attention in GLE for better feature fusion.
We are also exploring more efficient guiding mechanisms for future work.

\subsection{The Effectiveness of Guide Token}

We conducted comparative experiments to evaluate the effectiveness of the proposed $[Guide]$ token against two mainstream methods: the class token~\cite{dosovitskiy2020image} and global average pooling (GAP)~\cite{chu2021conditional}, as presented in Tab.~\ref{table:clstoken}.
The results demonstrate that the $[Guide]$ token improves model performance by approximately 1.7\% compared to these methods, without significantly increasing the number of parameters.

\begin{table}[H]
    \centering
    \resizebox{0.8\columnwidth}{!}{
    \begin{tabular}{c|c|c|c}
        \toprule
        Backbone                & Method                                         & \makecell[c]{\#P \\ (M)}         & \makecell[c]{Top-1 Acc. \\ (\%)} \\
        \midrule
        \multirow{3}*{PVT-Tiny} & GAP                                            & 14.0                             & 77.7 (-1.8)                      \\
        ~                       & Class token                                    & 14.0                             & 77.9 (-1.6)                      \\
        ~                       & \cellcolor{gray!20}\textbf{Guide token (ours)} & \cellcolor{gray!20}\textbf{14.0} & \cellcolor{gray!20}\textbf{79.5} \\
        \bottomrule
    \end{tabular}}
    \caption{Comparison between Guide token and other methods.}
    \label{table:clstoken}
\end{table}

Furthermore, an important observation from Tab.~\ref{table:clstoken} and Tab.~\ref{abla} is that when the local window size of self-attention and the kernel size of convolution are both set to $2 \times 2$, the self-attention method achieves higher accuracy than the convolution method.
This suggests that the self-attention mechanism has a superior capability in extracting high-frequency features.

\subsection{The Kernel Size of GTG}

To determine the optimal kernel size for GTG, we conducted a series of performance comparison experiments with different kernel sizes, as shown in Tab.~\ref{table:stgks}.

\begin{table}[H]
    \centering
    \resizebox{0.7\columnwidth}{!}{
    \begin{tabular}{c|c|c|c}
        \toprule
        Backbone                & Kernel Size                         & \makecell[c]{\#P \\ (M)}         & \makecell[c]{Top-1 Acc. \\ (\%)}  \\
        \midrule
        \multirow{6}*{PVT-Tiny} & $1\times 1$                         & 13.6                             & 78.17                             \\
        ~                       & $3\times 3$                         & 13.6                             & 78.86                             \\
        ~                       & $5\times 5$                         & 13.6                             & 79.43                             \\
        ~                       & \cellcolor{gray!20}$\bm{7\times 7}$ & \cellcolor{gray!20}\textbf{13.6} & \cellcolor{gray!20}\textbf{79.47} \\
        ~                       & $9\times 9$                         & 13.7                             & 79.46                             \\
        ~                       & $31\times 31$                       & 14.5                             & 79.34                             \\
        \bottomrule
    \end{tabular}}
    \caption{Performance comparison of GTG with different kernel sizes. “\#P” is calculated by the \textit{thop} package.}
    \label{table:stgks}
\end{table}

\begin{figure}[t]
    \centering
    \includegraphics[width=\columnwidth]{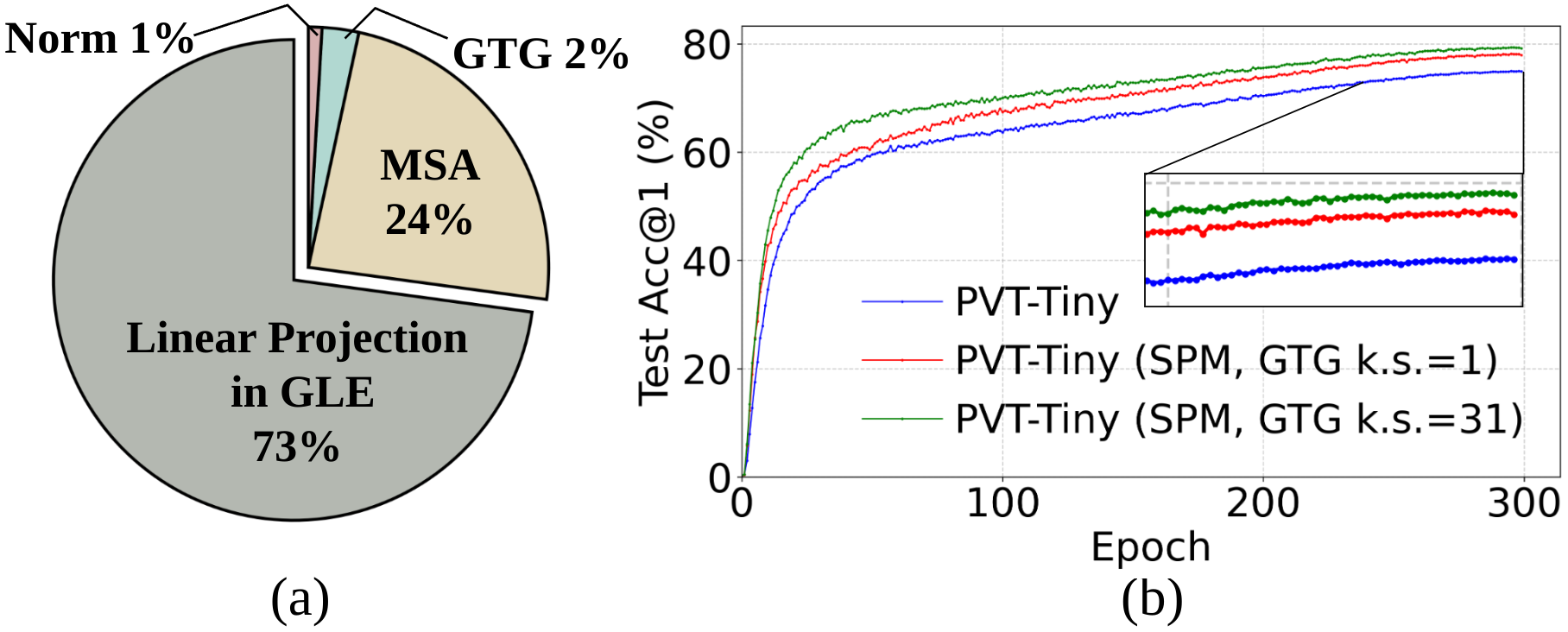}
    \caption{
    (a) FLOPs distribution across modules in SPM.
    (b) Stable convergence even with extreme kernel sizes (1 and 31).
    }
    \label{flops_loss}
\end{figure}

To assess optimization stability, we experimented with kernel sizes at extremes (1 and 31) and monitored training dynamics. As shown in Fig.~\ref{flops_loss}~(b), training remains stable and converges faster than the baseline (PVT-Tiny), indicating that our $[Guide]$ token design does not introduce instability.
Additionally, we had initially added residual connections to $[Guide]$ tokens, but found that removing them slightly improved performance, and the training remained stable throughout.

\subsection{Gradually Applying SPM}
From Tab.~\ref{table:replace}, we conclude that 1) SPM is robust to varying input sizes containing different levels of semantic content, consistently improving performance (2nd and 3rd rows), and 2) repeated application of SPM leads to a healthy linear increase in performance (2nd, 4th and 5th rows).

\begin{table}[H]
    \centering
    \resizebox{0.8\columnwidth}{!}{
    \begin{tabular}{c|ccc|c|c}
        \toprule
        Backbone                & Stage1     & Stage2     & Stage3     & \makecell[c]{\#P \\ (M)} & \makecell[c]{Top-1 Acc. \\ (\%)} \\
        \midrule
        \multirow{5}*{PVT-Tiny} &            &            &            & 13.2                     & 75.1                             \\
        ~                       & \checkmark &            &            & 13.3                     & 76.4 (+1.3)                      \\
        ~                       &            &            & \checkmark & 13.7                     & 76.7 (+1.6)                      \\
        ~                       & \checkmark & \checkmark &            & 13.5                     & 78.1 (+3.0)                      \\
        ~                       & \checkmark & \checkmark & \checkmark & 14.0                     & 79.5 (+4.4)                      \\
        \bottomrule
    \end{tabular}}
    \caption{Gradually replacing the original patch merging in PVT-Tiny with SPM.}
    \label{table:replace}
\end{table}

%% file: src/6_Future_Work.tex
\section{Future Work}

The current SPM involves many hyperparameters that need to be considered; future work will explore a more concise and computationally efficient stepwise structure to further enhance HVTs performance, and experiments demonstrate the effectiveness of the global-guide mechanism, which will be analyzed in greater depth.

%% file: src/7_Conclusion.tex
\section{Conclusion}

In this work, we introduced the Stepwise Patch Merging, inspired by the brain's ability to integrate global and local information for comprehensive visual understanding.
The proposed SPM framework, comprising Multi-Scale Aggregation and Guided Local Enhancement modules, demonstrates significant improvements in various computer vision tasks, including classification, detection, and segmentation.
Meanwhile, we showed that SPM consistently enhances the performance of different backbone models.
The robustness of SPM to different input sizes and its effective generalization further underscore its versatility and potential as a powerful enhancement technique.

%% file: ijcai25.bbl
\begin{thebibliography}{}

\bibitem[\protect\citeauthoryear{Bolya \bgroup \em et al.\egroup }{2022}]{bolya2022token}
Daniel Bolya, Cheng-Yang Fu, Xiaoliang Dai, Peizhao Zhang, Christoph Feichtenhofer, and Judy Hoffman.
\newblock Token merging: Your vit but faster.
\newblock {\em arXiv preprint arXiv:2210.09461}, 2022.

\bibitem[\protect\citeauthoryear{Chen \bgroup \em et al.\egroup }{2023}]{chen2023building}
Yongjie Chen, Hongmin Liu, Haoran Yin, and Bin Fan.
\newblock Building vision transformers with hierarchy aware feature aggregation.
\newblock In {\em Proceedings of the IEEE/CVF International Conference on Computer Vision}, pages 5908--5918, 2023.

\bibitem[\protect\citeauthoryear{Chu \bgroup \em et al.\egroup }{2021}]{chu2021conditional}
Xiangxiang Chu, Zhi Tian, Bo~Zhang, Xinlong Wang, Xiaolin Wei, Huaxia Xia, and Chunhua Shen.
\newblock Conditional positional encodings for vision transformers.
\newblock {\em arXiv preprint arXiv:2102.10882}, 2021.

\bibitem[\protect\citeauthoryear{Dai \bgroup \em et al.\egroup }{2017}]{dai2017deformable}
Jifeng Dai, Haozhi Qi, Yuwen Xiong, Yi~Li, Guodong Zhang, Han Hu, and Yichen Wei.
\newblock Deformable convolutional networks.
\newblock In {\em Proceedings of the IEEE international conference on computer vision}, pages 764--773, 2017.

\bibitem[\protect\citeauthoryear{Deng \bgroup \em et al.\egroup }{2009}]{deng2009imagenet}
Jia Deng, Wei Dong, Richard Socher, Li-Jia Li, Kai Li, and Li~Fei-Fei.
\newblock Imagenet: A large-scale hierarchical image database.
\newblock In {\em 2009 IEEE conference on computer vision and pattern recognition}, pages 248--255. Ieee, 2009.

\bibitem[\protect\citeauthoryear{Devlin \bgroup \em et al.\egroup }{2018}]{devlin2018bert}
Jacob Devlin, Ming-Wei Chang, Kenton Lee, and Kristina Toutanova.
\newblock Bert: Pre-training of deep bidirectional transformers for language understanding.
\newblock {\em arXiv preprint arXiv:1810.04805}, 2018.

\bibitem[\protect\citeauthoryear{Dong \bgroup \em et al.\egroup }{2022}]{dong2022cswin}
Xiaoyi Dong, Jianmin Bao, Dongdong Chen, Weiming Zhang, Nenghai Yu, Lu~Yuan, Dong Chen, and Baining Guo.
\newblock Cswin transformer: A general vision transformer backbone with cross-shaped windows.
\newblock In {\em Proceedings of the IEEE/CVF Conference on Computer Vision and Pattern Recognition}, pages 12124--12134, 2022.

\bibitem[\protect\citeauthoryear{Dosovitskiy \bgroup \em et al.\egroup }{2020}]{dosovitskiy2020image}
Alexey Dosovitskiy, Lucas Beyer, Alexander Kolesnikov, Dirk Weissenborn, Xiaohua Zhai, Thomas Unterthiner, Mostafa Dehghani, Matthias Minderer, Georg Heigold, Sylvain Gelly, et~al.
\newblock An image is worth 16x16 words: Transformers for image recognition at scale.
\newblock {\em arXiv preprint arXiv:2010.11929}, 2020.

\bibitem[\protect\citeauthoryear{Du \bgroup \em et al.\egroup }{2016}]{du2016study}
Min Du, Shili Ding, and Huancheng Jia.
\newblock Study on density peaks clustering based on k-nearest neighbors and principal component analysis.
\newblock {\em Knowledge-Based Systems}, 99:135--145, 2016.

\bibitem[\protect\citeauthoryear{Ebert \bgroup \em et al.\egroup }{2023}]{ebert2023plg}
Nikolas Ebert, Didier Stricker, and Oliver Wasenm{\"u}ller.
\newblock Plg-vit: Vision transformer with parallel local and global self-attention.
\newblock {\em Sensors}, 23(7):3447, 2023.

\bibitem[\protect\citeauthoryear{Gilbert and Li}{2013}]{gilbert2013top}
Charles~D Gilbert and Wu~Li.
\newblock Top-down influences on visual processing.
\newblock {\em Nature reviews neuroscience}, 14(5):350--363, 2013.

\bibitem[\protect\citeauthoryear{Glorot and Bengio}{2010}]{glorot2010understanding}
Xavier Glorot and Yoshua Bengio.
\newblock Understanding the difficulty of training deep feedforward neural networks.
\newblock In {\em Proceedings of the thirteenth international conference on artificial intelligence and statistics}, JMLR Workshop and Conference Proceedings, pages 249--256, 2010.

\bibitem[\protect\citeauthoryear{Guo \bgroup \em et al.\egroup }{2022}]{guo2022cmt}
Jianyuan Guo, Kai Han, Han Wu, Yehui Tang, Xinghao Chen, Yunhe Wang, and Chang Xu.
\newblock Cmt: Convolutional neural networks meet vision transformers.
\newblock In {\em Proceedings of the IEEE/CVF Conference on Computer Vision and Pattern Recognition}, pages 12175--12185, 2022.

\bibitem[\protect\citeauthoryear{Hassani \bgroup \em et al.\egroup }{2023}]{hassani2023neighborhood}
Ali Hassani, Steven Walton, Jiachen Li, Shen Li, and Humphrey Shi.
\newblock Neighborhood attention transformer.
\newblock In {\em Proceedings of the IEEE/CVF Conference on Computer Vision and Pattern Recognition}, pages 6185--6194, 2023.

\bibitem[\protect\citeauthoryear{He \bgroup \em et al.\egroup }{2016}]{he2016deep}
Kaiming He, Xiangyu Zhang, Shaoqing Ren, and Jian Sun.
\newblock Deep residual learning for image recognition.
\newblock In {\em Proceedings of the IEEE conference on computer vision and pattern recognition}, pages 770--778, 2016.

\bibitem[\protect\citeauthoryear{He \bgroup \em et al.\egroup }{2017}]{he2017mask}
Kaiming He, Georgia Gkioxari, Piotr Doll{\'a}r, et~al.
\newblock Mask r-cnn.
\newblock In {\em Proceedings of the IEEE International Conference on Computer Vision}, pages 2961--2969, 2017.

\bibitem[\protect\citeauthoryear{Hou \bgroup \em et al.\egroup }{2022}]{hou2022conv2former}
Qibin Hou, Cheng-Ze Lu, Ming-Ming Cheng, and Jiashi Feng.
\newblock Conv2former: A simple transformer-style convnet for visual recognition.
\newblock {\em arXiv preprint arXiv:2211.11943}, 2022.

\bibitem[\protect\citeauthoryear{Hubel and Wiesel}{1962}]{hubel1962receptive}
David~H Hubel and Torsten~N Wiesel.
\newblock Receptive fields, binocular interaction and functional architecture in the cat's visual cortex.
\newblock {\em The Journal of physiology}, 160(1):106, 1962.

\bibitem[\protect\citeauthoryear{Kirillov \bgroup \em et al.\egroup }{2019}]{kirillov2019panoptic}
Alexander Kirillov, Ross Girshick, Kaiming He, and Piotr Doll{\'a}r.
\newblock Panoptic feature pyramid networks.
\newblock In {\em Proceedings of the IEEE/CVF conference on computer vision and pattern recognition}, pages 6399--6408, 2019.

\bibitem[\protect\citeauthoryear{Krizhevsky \bgroup \em et al.\egroup }{2012}]{krizhevsky2012imagenet}
Alex Krizhevsky, Ilya Sutskever, and Geoffrey~E Hinton.
\newblock Imagenet classification with deep convolutional neural networks.
\newblock {\em Advances in neural information processing systems}, 25, 2012.

\bibitem[\protect\citeauthoryear{Li \bgroup \em et al.\egroup }{2019}]{li2019selective}
Xiang Li, Wenhai Wang, Xiaolin Hu, and Jian Yang.
\newblock Selective kernel networks.
\newblock In {\em Proceedings of the IEEE/CVF conference on computer vision and pattern recognition}, pages 510--519, 2019.

\bibitem[\protect\citeauthoryear{Liang \bgroup \em et al.\egroup }{2022}]{liang2022not}
Youwei Liang, Chongjian Ge, Zhan Tong, Yibing Song, Jue Wang, and Pengtao Xie.
\newblock Not all patches are what you need: Expediting vision transformers via token reorganizations.
\newblock {\em arXiv preprint arXiv:2202.07800}, 2022.

\bibitem[\protect\citeauthoryear{Lin \bgroup \em et al.\egroup }{2014}]{lin2014microsoft}
Tsung-Yi Lin, Michael Maire, Serge Belongie, et~al.
\newblock Microsoft coco: Common objects in context.
\newblock In {\em Computer Vision--ECCV 2014: 13th European Conference, Zurich, Switzerland, September 6-12, 2014, Proceedings, Part V}, volume~13, pages 740--755. Springer International Publishing, 2014.

\bibitem[\protect\citeauthoryear{Lin \bgroup \em et al.\egroup }{2023}]{lin2023scale}
Weifeng Lin, Ziheng Wu, Jiayu Chen, Jun Huang, and Lianwen Jin.
\newblock Scale-aware modulation meet transformer.
\newblock In {\em Proceedings of the IEEE/CVF international conference on computer vision}, pages 6015--6026, 2023.

\bibitem[\protect\citeauthoryear{Liu \bgroup \em et al.\egroup }{2021}]{liu2021swin}
Ze~Liu, Yutong Lin, Yue Cao, Han Hu, Yixuan Wei, Zheng Zhang, Stephen Lin, and Baining Guo.
\newblock Swin transformer: Hierarchical vision transformer using shifted windows.
\newblock In {\em Proceedings of the IEEE/CVF international conference on computer vision}, pages 10012--10022, 2021.

\bibitem[\protect\citeauthoryear{Loshchilov and Hutter}{2016}]{loshchilov2016sgdr}
Ilya Loshchilov and Frank Hutter.
\newblock Sgdr: Stochastic gradient descent with warm restarts.
\newblock {\em arXiv preprint arXiv:1608.03983}, 2016.

\bibitem[\protect\citeauthoryear{Loshchilov and Hutter}{2017}]{loshchilov2017decoupled}
Ilya Loshchilov and Frank Hutter.
\newblock Decoupled weight decay regularization.
\newblock {\em arXiv preprint arXiv:1711.05101}, 2017.

\bibitem[\protect\citeauthoryear{Luo \bgroup \em et al.\egroup }{2016}]{luo2016understanding}
Wenjie Luo, Yujia Li, Raquel Urtasun, and Richard Zemel.
\newblock Understanding the effective receptive field in deep convolutional neural networks.
\newblock {\em Advances in neural information processing systems}, 29, 2016.

\bibitem[\protect\citeauthoryear{Pan \bgroup \em et al.\egroup }{2022}]{pan2022less}
Zizheng Pan, Bohan Zhuang, Haoyu He, Jing Liu, and Jianfei Cai.
\newblock Less is more: Pay less attention in vision transformers.
\newblock In {\em Proceedings of the AAAI Conference on Artificial Intelligence}, volume~36, pages 2035--2043, 2022.

\bibitem[\protect\citeauthoryear{Rao \bgroup \em et al.\egroup }{2021}]{rao2021dynamicvit}
Yongming Rao, Wenliang Zhao, Benlin Liu, Jiwen Lu, Jie Zhou, and Cho-Jui Hsieh.
\newblock Dynamicvit: Efficient vision transformers with dynamic token sparsification.
\newblock {\em Advances in neural information processing systems}, 34:13937--13949, 2021.

\bibitem[\protect\citeauthoryear{Ren \bgroup \em et al.\egroup }{2022}]{ren2022shunted}
Sucheng Ren, Daquan Zhou, Shengfeng He, Jiashi Feng, and Xinchao Wang.
\newblock Shunted self-attention via multi-scale token aggregation.
\newblock In {\em Proceedings of the IEEE/CVF Conference on Computer Vision and Pattern Recognition}, pages 10853--10862, 2022.

\bibitem[\protect\citeauthoryear{Szegedy \bgroup \em et al.\egroup }{2015}]{szegedy2015going}
Christian Szegedy, Wei Liu, Yangqing Jia, Pierre Sermanet, Scott Reed, Dragomir Anguelov, Dumitru Erhan, Vincent Vanhoucke, and Andrew Rabinovich.
\newblock Going deeper with convolutions.
\newblock In {\em Proceedings of the IEEE conference on computer vision and pattern recognition}, pages 1--9, 2015.

\bibitem[\protect\citeauthoryear{Szegedy \bgroup \em et al.\egroup }{2016}]{szegedy2016rethinking}
Christian Szegedy, Vincent Vanhoucke, Sergey Ioffe, et~al.
\newblock Rethinking the inception architecture for computer vision.
\newblock In {\em Proceedings of the IEEE conference on computer vision and pattern recognition}, pages 2818--2826, 2016.

\bibitem[\protect\citeauthoryear{Touvron \bgroup \em et al.\egroup }{2021}]{touvron2021training}
Hugo Touvron, Matthieu Cord, Matthijs Douze, Francisco Massa, Alexandre Sablayrolles, and Herv{\'e} J{\'e}gou.
\newblock Training data-efficient image transformers \& distillation through attention.
\newblock In {\em International conference on machine learning}, pages 10347--10357. PMLR, 2021.

\bibitem[\protect\citeauthoryear{Vaswani \bgroup \em et al.\egroup }{2017}]{vaswani2017attention}
Ashish Vaswani, Noam Shazeer, Niki Parmar, Jakob Uszkoreit, Llion Jones, Aidan~N Gomez, {\L}ukasz Kaiser, and Illia Polosukhin.
\newblock Attention is all you need.
\newblock {\em Advances in neural information processing systems}, 30, 2017.

\bibitem[\protect\citeauthoryear{Wang \bgroup \em et al.\egroup }{2020}]{wang2020deep}
Jingdong Wang, Ke~Sun, Tianheng Cheng, Borui Jiang, Chaorui Deng, Yang Zhao, Dong Liu, Yadong Mu, Mingkui Tan, Xinggang Wang, et~al.
\newblock Deep high-resolution representation learning for visual recognition.
\newblock {\em IEEE transactions on pattern analysis and machine intelligence}, 43(10):3349--3364, 2020.

\bibitem[\protect\citeauthoryear{Wang \bgroup \em et al.\egroup }{2021}]{wang2021pyramid}
Wenhai Wang, Enze Xie, Xiang Li, Deng-Ping Fan, Kaitao Song, Ding Liang, Tong Lu, Ping Luo, and Ling Shao.
\newblock Pyramid vision transformer: A versatile backbone for dense prediction without convolutions.
\newblock In {\em Proceedings of the IEEE/CVF international conference on computer vision}, pages 568--578, 2021.

\bibitem[\protect\citeauthoryear{Xie \bgroup \em et al.\egroup }{2017}]{xie2017aggregated}
Saining Xie, Ross Girshick, Piotr Doll{\'a}r, Zhuowen Tu, and Kaiming He.
\newblock Aggregated residual transformations for deep neural networks.
\newblock In {\em Proceedings of the IEEE conference on computer vision and pattern recognition}, pages 1492--1500, 2017.

\bibitem[\protect\citeauthoryear{Yue \bgroup \em et al.\egroup }{2021}]{yue2021vision}
Xiaoyu Yue, Shuyang Sun, Zhanghui Kuang, Meng Wei, Philip~HS Torr, Wayne Zhang, and Dahua Lin.
\newblock Vision transformer with progressive sampling.
\newblock In {\em Proceedings of the IEEE/CVF International Conference on Computer Vision}, pages 387--396, 2021.

\bibitem[\protect\citeauthoryear{Yun \bgroup \em et al.\egroup }{2019}]{yun2019cutmix}
Sangdoo Yun, Dongyoon Han, Seong~Joon Oh, Sanghyuk Chun, Junsuk Choe, and Youngjoon Yoo.
\newblock Cutmix: Regularization strategy to train strong classifiers with localizable features.
\newblock In {\em Proceedings of the IEEE/CVF international conference on computer vision}, pages 6023--6032, 2019.

\bibitem[\protect\citeauthoryear{Zeng \bgroup \em et al.\egroup }{2022}]{zeng2022not}
Wang Zeng, Sheng Jin, Wentao Liu, Chen Qian, Ping Luo, Wanli Ouyang, and Xiaogang Wang.
\newblock Not all tokens are equal: Human-centric visual analysis via token clustering transformer.
\newblock In {\em Proceedings of the IEEE/CVF Conference on Computer Vision and Pattern Recognition}, pages 11101--11111, 2022.

\bibitem[\protect\citeauthoryear{Zhang \bgroup \em et al.\egroup }{2017}]{zhang2017mixup}
Hongyi Zhang, Moustapha Cisse, Yann~N Dauphin, and David Lopez-Paz.
\newblock mixup: Beyond empirical risk minimization.
\newblock {\em arXiv preprint arXiv:1710.09412}, 2017.

\bibitem[\protect\citeauthoryear{Zhang \bgroup \em et al.\egroup }{2018}]{zhang2018shufflenet}
Xiangyu Zhang, Xinyu Zhou, Mengxiao Lin, and Jian Sun.
\newblock Shufflenet: An extremely efficient convolutional neural network for mobile devices.
\newblock In {\em Proceedings of the IEEE conference on computer vision and pattern recognition}, pages 6848--6856, 2018.

\bibitem[\protect\citeauthoryear{Zhong \bgroup \em et al.\egroup }{2020}]{zhong2020random}
Zhun Zhong, Liang Zheng, Guoliang Kang, Shaozi Li, and Yi~Yang.
\newblock Random erasing data augmentation.
\newblock In {\em Proceedings of the AAAI conference on artificial intelligence}, volume~34, pages 13001--13008, 2020.

\bibitem[\protect\citeauthoryear{Zhou \bgroup \em et al.\egroup }{2017}]{zhou2017scene}
Bolei Zhou, Hang Zhao, Xavier Puig, Sanja Fidler, Adela Barriuso, and Antonio Torralba.
\newblock Scene parsing through ade20k dataset.
\newblock In {\em Proceedings of the IEEE conference on computer vision and pattern recognition}, pages 633--641, 2017.

\end{thebibliography}
